%% file: main_ArXiv.tex
\documentclass{ecai} 

\usepackage{latexsym}
\usepackage{amssymb}
\usepackage{amsmath}
\usepackage{amsthm}
\usepackage{booktabs}
\usepackage{enumitem}
\usepackage{graphicx}
\usepackage{color}

\usepackage{subcaption}
\usepackage{multirow}
\usepackage{float}
\usepackage{caption}

\input{packages}

\newcommand{\BibTeX}{B\kern-.05em{\sc i\kern-.025em b}\kern-.08em\TeX}

\input{macros}

\begin{document}


\begin{frontmatter}


\paperid{\todo{123}} 


\title{Explainable Prediction of the Mechanical Properties of \\ Composites 
with CNNs 
} 


\author[A]{\fnms{Varun}~\snm{Raaghav}
}
\author[A]{\fnms{Dimitrios}~\snm{Bikos}
\footnote{Corresponding Authors. Emails: d.bikos17@imperial.ac.uk, antonio.rago@kcl.ac.uk.}
}
\author[B,C]{\fnms{Antonio}~\snm{Rago}
\footnotemark
}
\author[B]{\fnms{Francesca}~\snm{Toni}
}
\author[A]{\fnms{Maria}~\snm{Charalambides}
}

\address[A]{Department of Mechanical Engineering, Imperial College London, UK}
\address[B]{Department of Computing, Imperial College London, UK}
\address[C]{Department of Informatics, King's College London, UK}


\begin{abstract}
Composites are 
amongst the most important materials 
manufactured today, as evidenced by their use in countless applications. 
In order to establish the suitability of composites in specific applications,
finite element (FE) modelling, a numerical method based on partial differential equations, is the industry standard for assessing their mechanical properties.
However, FE modelling is exceptionally costly from a computational viewpoint, a limitation which has led to efforts 
towards applying AI models to this task. 
However, in these approaches: the chosen model architectures were rudimentary, feed-forward neural networks giving limited accuracy; the studies focus\DB{ed} on predicting elastic mechanical properties, without considering material strength limits; 
and the models lacked transparency, hindering trustworthiness by users.
In this paper, we show that convolutional neural networks (CNNs) equipped with methods from explainable AI (XAI) can be successfully deployed to solve this problem. 
Our approach uses customised CNNs trained on a dataset we generate using 
transverse tension tests 
in FE modelling to predict composites' mechanical properties, i.e., Young’s modulus and yield strength.
We show empirically that our approach achieves high accuracy, outperforming a baseline, ResNet-34, in estimating the mechanical properties.
We then use SHAP and Integrated Gradients, two post-hoc XAI methods, to explain the predictions, showing that the CNNs use the critical geometrical features that influence the composites' behaviour, thus allowing engineers to verify that the models are trustworthy by representing the science of composites.

\end{abstract}

\end{frontmatter}


\section{Introduction}
\label{sec:introduction}

Composites are one of the most important materials that are manufactured today due to their exceptional specific mechanical properties, i.e., mechanical properties normalised by density. 
This has resulted in their widespread application, e.g. in wind turbines, aerospace components and sports equipment \cite{rajak2021}.
The evaluation of these mechanical properties, and thus the performance and suitability of composites in individual applications, can be assessed via a variety of analytical, experimental and numerical methods \cite{he2015,varandas2020}. 
Analytical models provide accurate estimations for simplistic cases but rely on idealised assumptions (e.g. uniform inclusion positions and geometries) and do not account for all necessary geometrical or material parameters (e.g. weak bonding and complex constitutive laws) \cite{Ford_21}.
Experimental methods can be challenging and time-consuming \cite{harper1993}, especially when analysing how and why a material or system fails. Further\VR{more}, post-failure analysis often requires multiple steps (e.g. cutting samples and using optical tools), while in-situ testing demands specialised and costly equipment (e.g. in-situ testing rigs, X-ray tomography equipment, or high-speed cameras) \cite{wang2019}. These requirements can make experimental methods less practical for studies that explore how different factors influence failure behaviour.
Numerical models, e.g. employing finite element (FE) modelling, can provide accurate estimations of composites' properties and incorporate more realistic assumptions about their geometries, unlike analytical models. However, they are limited to small, simulated volumes containing only a few inclusions due to their extensive computational cost \cite{pathan2019} 
(in some cases simulations required up to 13 days to complete \cite{varandas2020}). Attempts have been made within the composite research community to develop more efficient numerical models, but these are often tailored to specific loading scenarios and types of composites and require prior knowledge of the key underlying physical mechanisms governing the composite behaviour \cite{malgioglio2021,bikos2025}.
Thus, while numerical models are the favoured option for assessing the properties of composites, they are limited by these weaknesses.

Concurrently, the inherent versatility of AI models in handling various types of data (e.g. tabular or images), along with their computational efficiency (especially after a one-off training cost), has been demonstrated by their widespread deployment in engineering problems \cite{qiu2022,saquib2024,wang2024}. 
In composite engineering, AI models have recently been applied to various 
tasks, including property prediction~\cite{Ford_21},  sensitivity analysis of failure processes \cite{post2023,wan2023}, optimisation of composite properties \cite{abueidda2019} and manufacturing design \cite{qiu2022}. Some efforts have been made to apply AI models to predicting mechanical properties in composites \cite{Ford_21,pathan2019,zhou2021}. However, these studies either focus solely on predicting elastic properties \cite{Ford_21} or, when extending to strength limits, are constrained by limited training data (approximately 1,000 datapoints) and small composite microstructures (fewer than 15 fibres) \cite{pathan2019, zhou2021}. These limitations do not meet the criteria for statistically representative composite microstructures \cite{samaras2024}, raising concerns about the reliability of the results. 
Furthermore, most of the deployed AI model architectures, primarily feed-forward neural networks, remain relatively rudimentary, resulting in limited accuracy \cite{Ford_21, pathan2019}.
Finally, the\DB{se} AI models were not supplemented with explanations to allow engineers to verify that the patterns being learnt are indeed representing the science of composites, thus missing a crucial opportunity for fostering trust in the engineers.

\begin{figure*}[t]
    \centering
    \includegraphics[width=0.9\textwidth]{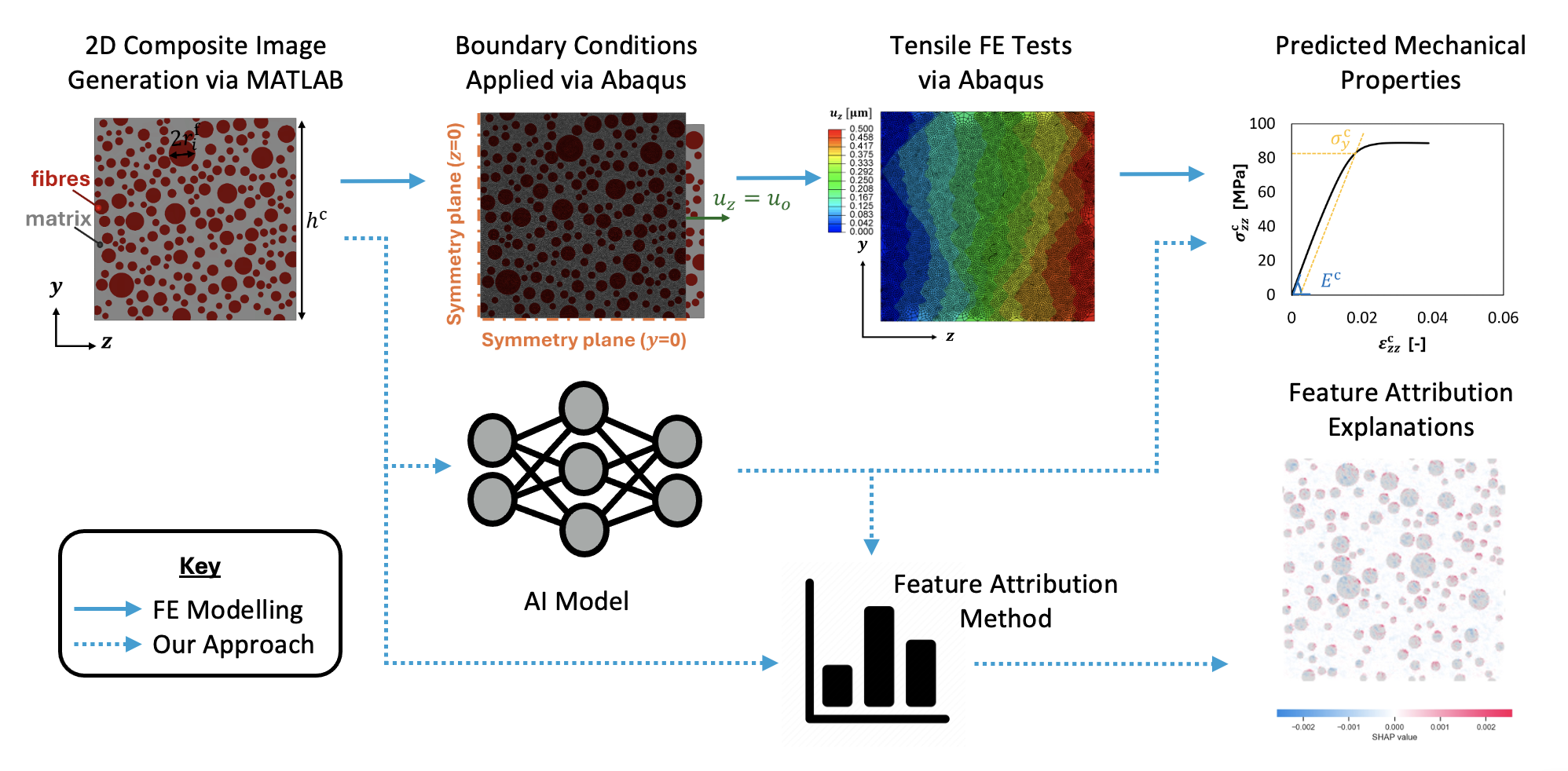}
    \caption{Comparison between FE modelling (the industry standard in mechanical engineering) and our approach, in which we alleviate the need for performing tensile tests, instead predicting the mechanical properties using CNNs supplemented with feature attribution explanations. 
    }
    \label{fig:pipeline}
\end{figure*}

In this paper, we propose a solution to this problem by adapting convolutional neural networks (CNNs)~\cite{Li_22}, supplemented with techniques from explainable AI (XAI, see~\cite{Ali_23} for a recent overview), to this image regression task.
To ensure that we have a representative training dataset for our CNN model and that we overcome the reported computational cost associated with 3D FE modelling of composites, we propose an FE-based framework for 2D composite microstructures that contain at least 100 fibres (a criterion for representative composite microstructures \cite{samaras2024}), as illustrated in Figure \ref{fig:pipeline}, to predict composite properties under transverse tension virtual tests.
We generate a database of 
2D composite microstructures, where the constituents exist in different proportions and spatial arrangements of fibres
, along with estimates of their mechanical properties (Young’s modulus and yield strength), obtained through FE-based tensile tests.
We then deploy XAI methods based on feature attribution, to explain the predictions, thus showing that the CNNs use the critical geometrical features that influence the composites' behaviour.
In doing so, we demonstrate that not only does our methodology produce predictions which
align closely with FE modelling in this budding research area, it also allows engineers to verify that the CNNs are representing the science of composites, thus increasing its trustworthiness.

After covering the related work (§\ref{sec:related}), we make the following contributions:
\begin{itemize}
    \item We generate a dataset of 7,893 images of (fibre-reinforced polymer matrix) composite microstructures and their mechanical properties (Young’s modulus and yield strength) under transverse tension tests via FE modelling, providing the code for doing so (§\ref{sec:dataset});
    \item We define a novel methodology for predicting the mechanical properties of composites from their microstructure images using customised CNNs with an optimised architecture (§\ref{ssec:CNN});
    \item We perform a 
    comparative analysis of two CNNs in predicting the mechanical properties of composites using our dataset (§\ref{ssec:AIresults}), showing that our customised architecture makes predictions closely correlated with FE modelling, outperforming ResNet-34 \cite{He_16} (adapted for this image regression task);
    \item We deploy XAI methods based on feature attribution (§\ref{sec:xai}), demonstrating 
    how the attributions values correlate with the FE modelling and thus identify the critical geometrical features that influence the composites’ behaviour in transverse tension.
\end{itemize}
Finally we conclude and look ahead to future work (§\ref{sec:conclusions}).

\section{Related Work}
\label{sec:related}


Over the last five years, significant efforts have been made in composite engineering to apply AI models, which have shown promising potential in property prediction, failure initiation identification, and microstructure reconstruction.	Here, we review each of these application areas in turn.

\textbf{Property Prediction. \;} 
Several studies have employed 2D FE modelling at the microscale to explicitly simulate the geometry of composite constituents (i.e., fibres, which provide the structural integrity of the composite, and a matrix, which holds the fibres in place) and investigate how variations in the spatial arrangement and proportion of fibres affect the composite's performance under transverse loading. These studies have analysed fibres with irregular shapes \cite{Ford_21}, stochastic geometries (e.g. checkerboard arrangements \cite{abueidda2019}), and spherical shapes \cite{Ford_21,pathan2019,zhou2021}. The outputs from these simulations served as training data for AI models, including artificial neural networks (ANNs, i.e., feed-forward multi-layer perceptrons), and CNNs, demonstrating high accuracy compared to FE simulations (treated as the ground truth here). However, these studies exhibit certain limitations. Some focus solely on predicting elastic moduli \cite{Ford_21}, properties that can already be accurately estimated using analytical models. Others require additional pre-processing steps, such as principal component analysis \cite{Ford_21,pathan2019}, before 2D composite microstructures can be used in AI models, while others utilise a small, non-representative training dataset \cite{pathan2019,zhou2021}. 

\textbf{Failure Initiation Identification. \;} 
Beyond property prediction, AI models have been applied to predict failure responses in composites by constructing full failure envelope maps, achieving at least 97\% accuracy \cite{chen2021, post2023, wan2023, yiben2024}. However, again many of these studies due to the high computational cost associated with FE modelling rely on a small, non-representative training dataset, limiting their broader applicability and accuracy.
For instance, \citet{chen2021} employed microscale FE modelling to predict composite failure responses, using results from 560 composite microstructures as training data for a CNN (an order of magnitude less than we consider here).
Similarly, \citet{wan2023} developed a high-fidelity FE-based approach with just six composite microstructures to estimate failure points under different loading conditions. These results were then used as training data for two ANN models, but each FE simulation required up to 72 hours for completion.
Meanwhile, \citet{post2023} proposed an analytical algorithm to estimate damage initiation in composites based on arbitrary 3D stress states. The output from this failure model served as training data for various AI models, although the training dataset size was not specified. These models were evaluated on 10,000 test samples, achieving up to 99\% accuracy.

\textbf{Microstructure Reconstruction. \;} 
Reversing the microstructure-property relationship, studies have developed AI models to estimate the fibre orientation, a key parameter directly linked to a composite's strength, in materials with discontinuous fibres \cite{larson2022,saquib2024}. More specifically, these studies obtained strain fields, mappings that represent the degree of composite deformation under testing, in different directions from FE modelling and used them to predict the average fibre orientation tensor across the composite thickness. The resulting data were then used to train an ANN, which demonstrated strong predictive capabilities.

A key difference between these studies and our approach is that none of the above provided explanations of the AI models or transparency in their tuning parameters. As a result, AI models in this field often function as "black boxes," requiring extensive validation to ensure predictive reliability. This limitation reduces AI applications in composites to a curve-fitting exercise rather than an interpretable modelling approach, posing a significant disadvantage compared to numerical methods. We posit that this challenge can be addressed through XAI methods.

\textbf{XAI in Engineering. \;} 
XAI can enhance interpretability by providing insights into AI model decisions and uncovering information that finite element models alone may not readily offer. Integrating AI with FE modelling in a synergistic way can enhance our understanding of the geometrical mechanisms influencing composite performance. However, despite its potential, XAI has not yet been widely applied in the composite engineering field. A key example demonstrating the importance of XAI in composite applications is the work by \citet{yossef2024} which introduced a powerful AI framework integrating FE modelling and machine learning models.
In their study, FE modelling was used to predict composite strength (i.e., the FE modelling output) based on the material properties of laminates with different ply stacking sequences (i.e., the FE modelling input). A dataset of 1,000 composite strength values from the FE modelling was then employed to train an AI model (XGBoost \cite{Chen_16}). XGBoost was then coupled with XAI techniques (including SHAP \cite{Lundberg_17}) to identify the material properties with the greatest influence on composite strength, suggest an optimal laminate layup, and demonstrate the potential of XAI for advancing scientific discovery in this domain.

To the best of our knowledge, our approach is the first to address explainability challenges in the property prediction of composites using AI models to understand the contribution of microscopic geometrical parameters to performance.


\section{Generation of the Dataset}
\label{sec:dataset}

In this section we detail how we generated a dataset for mechanical property prediction.\footnote{The images and code for reproducing this dataset can be found at \url{https://doi.org/10.5281/zenodo.15343792}}. We first show how we generated the set of images (§\ref{ssec:images}); before using FE modelling to obtain the property values (§\ref{ssec:FEM}); and finally our evaluation of the property values against the volume fraction in the images (§\ref{ssec:FEresults}).

To ensure clarity in distinguishing measurements associated with different materials (i.e., the composite, its fibres and its matrix), superscripts are used throughout the text: $c$ denotes the composite, $f$ the fibres, and $m$ the matrix. This convention applies to all dimensional and material property values in this paper.

\subsection{Microstructure Image Generation}
\label{ssec:images}

Our first task was to synthesise a dataset of images of composite materials as inputs to the CNN.
To do so, artificial microstructures containing randomly positioned fibre centres were generated for different fibre volume fractions using the algorithm introduced by \citet{tschopp2008}. The algorithm, based on the random sequential adsorption (RSA) technique, enables control over the number, shape, and size of inclusions (in this case, circular fibres) within the 2D defined square boundary. 

A lognormal distribution was used to generate circular fibre radii, with a target mean radius of $r^f = 0.516 \mu m$. The lognormal distribution was defined in logarithmic space with a mean $\mu$ = $r^f /10 $ and standard deviation $S = r^f/20$. The resulting distribution was then uniformly scaled to achieve the target mean radius $r^f$ and to satisfy the specified fibre volume fraction (i.e., the proportion of the 2D boundary represented by fibres), $V^{f}$, as in \cite{tschopp2008}. The fibres were then randomly distributed within a 2D boundary, the matrix, to create a representation of the composite, as seen on the left of the top row of Figure \ref{fig:pipeline}.
The size of the 2D square-shaped boundary was controlled to be $h^{c} \times h^{c} = 25.8 \mu m \times 25.8 \mu m$), ensuring that at least 100 fibres are enclosed in it for all tested cases. This satisfied the criterion $h^{c}/r^f > 50$, ensuring statistically representative microstructures and converged composite properties \cite{samaras2024}.  

A total of 7,893 2D composite microstructures 
were generated\footnote{Only 7,893 out of 13,000 generated 2D microstructures were analysed. The remaining cases were excluded due to FE meshing failures, primarily from closely spaced fibres, and occasional disruptions in job scheduling on the high-performance computing system, which led to simulations with errors. 
}, with the fibre volume fraction, $V^{f}$, ranging from 20.0\% to 44.0\% in discrete intervals of 2.0\%. Although the fibre volume fraction in composites can reach up to $V^{f}$= 60.0\%, microstructures with $V^{f}$ > 44.0\% were not considered due to limitations in the generation algorithm, a known constraint of RSA techniques. 
The 2D composite image on the left of the top row of Figure \ref{fig:pipeline} shows a typical microstructure with a fibre volume fraction of 40.0\%.

\subsection{Calculation of the Mechanical Properties via FE Modelling}
\label{ssec:FEM}



Our next task was to import the composite microstructure images into the FE modelling software, generate a mesh of the geometry and define suitable numerical methods and boundary conditions to simulate a virtual tensile test. 
To create the 2D composite microstructure in Abaqus \cite{Abaqus}, details obtained from the RSA algorithm were used \cite{tschopp2008}, including the coordinates of the fibre centres and distribution of fibre radii. The 2D composite geometry was then meshed using linear quadrilateral plane strain elements (using the CPE4 element type in Abaqus), with a mean element size of approximately $1.0 \times r^f$, ensuring converged effective mechanical properties of the composite based on preliminary mesh convergence studies (these preliminary results are not shown in this paper). 

To simulate a virtual tensile test of the composite microstructure along the z-direction, we applied specific boundary conditions in the FE modelling. The boundary conditions included an arbitrary constant displacement ($u_z=+u_o$) at the composite's microstructure right edge ($z=h^{c}$) to simulate tension, while the opposite edge ($z=0$) was fixed in the z-direction ($u_z=0$). To prevent any undesired global (rigid body) motion of the microstructure, the bottom horizontal edge ($y=0$) was constrained in the y-direction ($u_y=0$). 
This is shown in the image on the second from the left of the top row of Figure \ref{fig:pipeline}, illustrating the mesh and applied boundary conditions. 

\begin{figure}[t]
    \centering
    \includegraphics[width=0.25\textwidth]{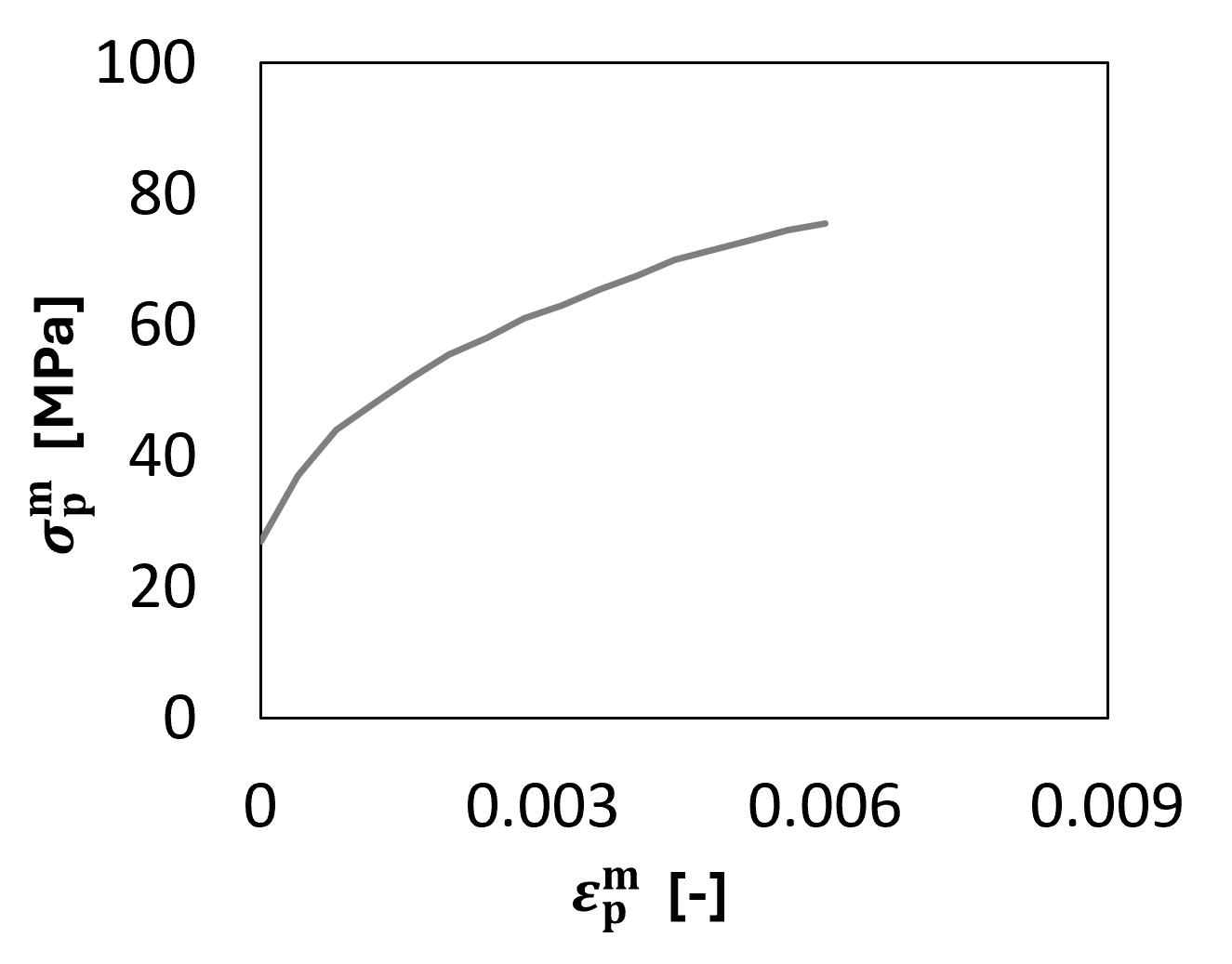}
    \caption{Plastic stress ($\sigma_{p}^m$) versus plastic strain ($\epsilon_{p}^m$) relationship of the F3900 epoxy matrix used in the FE modelling \cite{joshua_robbins_building_2019}.} 
    \label{fig:1c}
\end{figure}

In composite materials, the fibres primarily carry the load, while the matrix material holds the fibres in place and transfers stress between them. The fibres are commonly modelled as linear elastic materials with no plastic deformations involved. On the other hand, the matrix is often seen as a material with complex nonlinear material behaviour. 
In this work, the matrix was modelled as a nonlinear elastic-plastic material representing an F3900 epoxy, with Young’s modulus and Poisson’s ratio obtained experimentally in tension \cite{Ford_21,joshua_robbins_building_2019}. The matrix’s plastic stress-strain behaviour was implemented in Abaqus using the Von Mises yield criterion, a commonly used method for simulating plastic deformation in isotropic materials. 
Fibres were modelled as isotropic linear elastic, representing T800S fibres, with properties corresponding to their transverse direction \cite{Ford_21}. 
Table \ref{tab:material_constants} 
summarises the elastic properties of the matrix and fibres, 
while Figure \ref{fig:1c} shows the experimentally obtained plastic stress–strain data for the matrix used in the FE modelling.

To simulate quasi-static conditions under tension (i.e., no inertia/dynamic effects assumed), all composite microstructures were modelled using the implicit static solver in Abaqus 2022 \cite{Abaqus}. A high-performance computing system with 16 cores and a maximum of 8 GB of memory on an AMD EPYC 7742 server was utilised to run the FE simulations
. The time required for each FE simulation to capture the complete composite stress-strain behaviour, i.e., $\sigma_{zz}^c$ to reach a plateau, was approximately 2 minutes. Consequently, the overall approximate time to complete all 7,893 FE simulations was around 263 hours.


\begin{table}[t]
    \centering
    \begin{tabular}{ccc}
    \hline
    \textbf{Property} &
    \textbf{F3900 epoxy} &
    \textbf{T800 fibre} \\
    \hline
    Young's Modulus (E) [GPa] & 
    2.82 &
    15.51 \\
    Poisson's Ratio ($\nu$) & 
    0.387 &
    0.250 \\
    \hline
    \end{tabular}
    \caption{Elastic material properties of T800S fibre and F3900 epoxy matrix used in the FE modelling, in line with \cite{Ford_21,joshua_robbins_building_2019}.} 
    \label{tab:material_constants}
\end{table}

The final task of the FE modelling was to compute the bulk mechanical properties of the composite, specifically the elastic modulus ($E^c$) and yield strength ($\sigma_y^c$). These values were later used as 
the predicted variables in the training data for the CNNs.
The 
FE modelling provided outputs in the form of reaction forces ($F_{zz}^c$) and displacements in the z-direction ($u_{zz}^c$) at the right vertical edge of the microstructure (z = $h^c$), as shown in the image on the second from the right of the top row of Figure \ref{fig:pipeline}. These outputs were then converted into stress and strain values representative of the composite’s macroscopic behaviour. The stress, $\sigma_{zz}^c$, was calculated by dividing the reaction force at the right edge ($F_{zz}^c$) by the original cross-sectional area of the composite, defined as $A^c = h^c$ × (unit thickness). The corresponding strain, $\epsilon_{zz}^c$, was determined from the displacement of the right edge ($u_{zz}^c$) divided by the initial length of the composite microstructure, $L^c = h^c$.
The stress–strain relationship could then be plotted as in the image on the right of the top row of Figure \ref{fig:pipeline}.

The Young’s modulus of the composite ($E^c$) was determined as the gradient of the $\sigma_{zz}^c$ - $\epsilon_{zz}^c$ curve at the first increment of the simulation. The composite yield strength ($\sigma_y^c$) was approximated using the 0.2\% proof stress convention. The image on the right of the top row of Figure \ref{fig:pipeline} illustrates a typical example, highlighting the locations of $E^c$ and $\sigma_y^c$ measurements for a 2D composite microstructure with a $V^{f}=40.0\%$. The outputs for $E^c$ and $\sigma_y^c$ from the 7,893 FE modelling simulations were subsequently used for the CNN prediction (§\ref{sec:AI}) and application of XAI techniques (§\ref{sec:xai}).

\subsection{Evaluation of the Mechanical Properties}
\label{ssec:FEresults}

\begin{figure}[t]
    \centering
    \begin{subfigure}{0.4\textwidth}
        \centering
        \includegraphics[width=\textwidth]{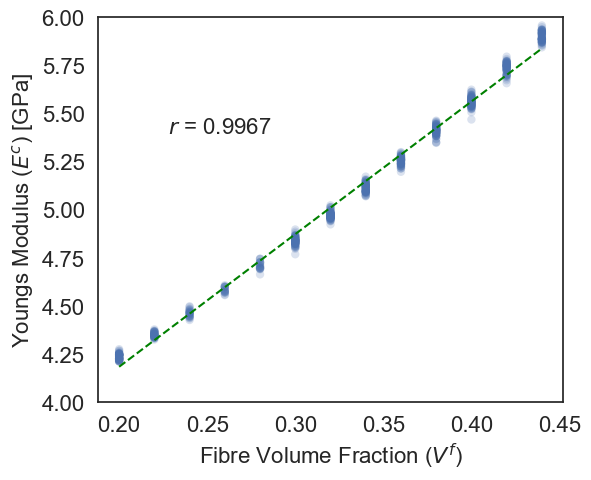}
        \caption{Young's modulus $E^{c}$ vs. fibre volume fraction $V^{f}$.}
        \label{fig:}
    \end{subfigure}
    \hspace{0.01\textwidth}
    \begin{subfigure}{0.4\textwidth}
        \centering
        \includegraphics[width=\textwidth]{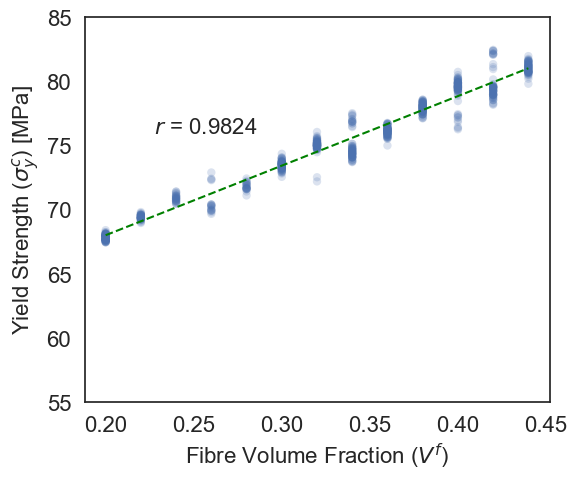}
        \caption{Yield strength $\sigma_y^c$ vs. the fibre volume fraction $V^{f}$.}
        \label{fig:}
    \end{subfigure}
    %
    \caption{FE modelling results summarising the relationship between the composite’s mechanical properties and fibre volume fraction. The green dashed lines in both plots represent linear trends, with the corresponding r values quantifying the linear correlation.}
    \label{fig:abaqus_results}
\end{figure}

Figure \ref{fig:abaqus_results} summarises the results of the FE modelling of the 7,893 composite microstructures in terms of the Young's modulus and the yield strength of the composite as a function of the fibre volume fraction. Both plots indicate a linear relationship between these two material properties and the fibre volume fraction, as evidenced by the high Pearson correlation factor, $r$. This linear relationship between the composite's material properties and fibre volume fraction is well established and widely reported in the composite engineering field for fibre volume fraction values less than 50\% \cite{raju2018}. While $V^{f}$ is widely reported as the dominant influencing factor in $E^c$ predictions \cite{Ford_21,pathan2019,raju2018}, the slightly greater variability observed in the $\sigma_y^c$ vs.\ $V^{f}$ trend suggests that $\sigma_y^c$ may be influenced by additional geometrical parameters, such as local fibre arrangements \cite{pathan2019} and interfibre distances.
The XAI analysis presented in §\ref{sec:xai} will shed further light on the contribution of the spatial fibre arrangements to materials properties of the composite
.

\section{Mechanical Property Prediction with CNNs}
\label{sec:AI}

We now show our methodology for predicting the mechanical properties with CNNs (§\ref{ssec:CNN}) and evaluation thereof (§\ref{ssec:AIresults}).

\subsection{Model Architectures and Training}
\label{ssec:CNN}

We employed two different CNNs using TensorFlow (v2.13.0) to determine if patterns present in the virtual microstructures could be identified and associated with the mechanical properties $E^c$ and $\sigma_y^c$. 
The CNNs were trained on the dataset created in §\ref{sec:dataset}, meaning they were deployed for the unorthodox task of image regression, adding to the novelty of this application of CNNs. 
One of each of the two CNNs was trained for each mechanical property, i.e., four CNNs were trained in total. We used an 85/10/5 split for training, testing and validation, respectively, with 5-fold cross-validation and trained for 10 epochs.
Hyperparameter tuning was performed during preliminary testing on the 5\% validation dataset.

\begin{figure}[t]
    \centering
    \includegraphics[width = 0.5\textwidth]{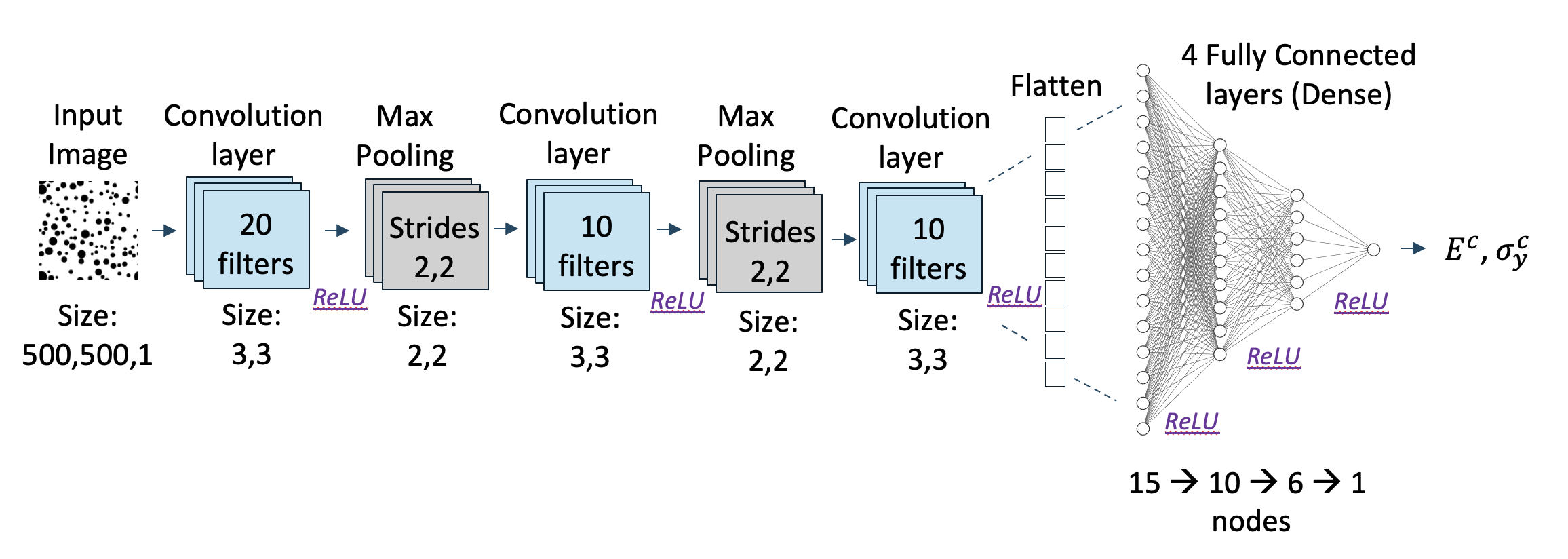}
    \caption{Our customised CNN's architecture, with microstructure images as inputs and mechanical properties ($E^c$ and $\sigma_y^c$) as outputs.}
    \label{fig:cnn_architecture}
\end{figure}

Our baseline architecture was 
ResNet-34, given its excellent performance history \cite{He_16}, modified such that the final layer adjusted to contain one node to allow for a regression output.
We then implemented our customised CNN of the architecture shown in Figure~\ref{fig:cnn_architecture}, which was the result of extensive hyperparameter optimisation.
(We leave the testing of other configurations of CNNs to future work.)

To assess our customised CNN against the baseline ResNet-34 architecture, we used a sample dataset which constituted 10\% of the full dataset due to limitations on computational cost and time, again using a 85/10/5 split and training for 10 epochs. 

\subsection{Prediction Results}
\label{ssec:AIresults}






\begin{table}[t]
    \centering
    \begin{tabular}{cccccc}
    \hline
    \multirow{2}{*}{\textbf{Dataset}} &
    \multirow{2}{*}{\textbf{Architecture}} &
    \multicolumn{2}{c}{\textbf{$E^{c}$}} &
    \multicolumn{2}{c}{\textbf{$\sigma_y^{c}$}} \\
     &
     &
    \textbf{$R^2$} &
    \!\!\textbf{Time}\!\! &
    \textbf{$R^2$} &
    \!\!\textbf{Time}\!\! \\
    \hline
        %
    %
    %
    \!Sample (10\%)\! &
    ResNet-34 & 
    \!$-1.2205$\! &
    $58$ &
    \!$-3.3193$\! &
    $61$ \\
    \!Sample (10\%)\! &
    \!Customised CNN\! & 
    $0.9583$ &
    $5$ &
    $0.9319$ &
    $5$\\
    Full (100\%) &
    \!Customised CNN\! & 
    $0.9852$ &
    $59$ &
    $0.9416$ &
    $50$ \\
    \hline
    \end{tabular}
    \caption{Correlation and training time (in minutes) for the CNNs' predictions of the mechanical properties performed on the sample ($10\%$) and full (from §\ref{sec:dataset}) datasets.} 
    \label{tab:performance_on_sample_dataset}
\end{table}

Table \ref{tab:performance_on_sample_dataset} shows our results for the CNNs in the form of their correlation with FE modelling and training time in predicting the mechanical properties on the sample and full datasets. 
It can be seen that the ResNet-34 architecture did not perform well in predicting the mechanical properties, with $R^{2}$ values of $-1.2205$ for $E^c$ and $-3.3193$ for $\sigma_y^c$. We believe that ResNet-34 would have likely achieved significantly better correlation with more training, but this architecture is somewhat overkill for this relatively simple task, given that our (much simpler) customised CNN performs well even when trained on the sample dataset. This is backed up by the training times (also shown in Table \ref{tab:performance_on_sample_dataset}), with our CNN being an order of magnitude faster. 

\begin{figure}[t]
    \centering
    \subfloat[\centering Customised CNN vs. FE modelling prediction of $E^{c}$. \label{subfig:const_YS}]{\includegraphics[width = 0.45\textwidth]{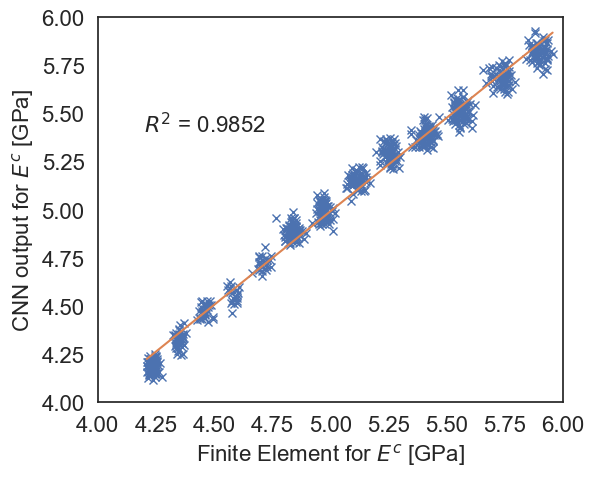}}
    \qquad
    \subfloat[\centering Customised CNN vs. FE modelling prediction of $\sigma_{y}^{c}$. \label{subfig:diff_YS}]{{\includegraphics[width=0.45\textwidth]{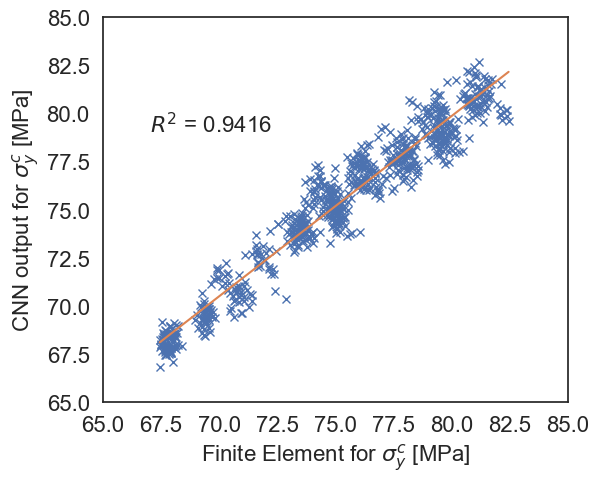} }}
    \caption{Customised CNN performance in predicting Young's modulus (a) and yield strength (b) of the composite. The orange lines in the two plots represent linear trends along with their associated $R^{2}$ values quantifying the correlation between the FE modelling and CNN approaches.} 
    \label{fig:performance}
\end{figure}

The results for the customised CNNs on the full dataset are  shown in Figure \ref{fig:performance}, indicating $R^2$ values of $0.9852$ for $E^c$ and $0.9416$ for $\sigma_y^c$. This level of accuracy is considerably higher than that achieved by recently reported AI models (e.g. $0.964$ for $E^c$ and $0.874$ for $\sigma_y^c$ using XGBoost \cite{pathan2019}).
The discrepancy in performance between predicting $E^c$ and $\sigma_y^c$ was also observed in the same study \cite{pathan2019}. The authors attributed this to the stronger dependence of $E^c$ on $V^{f}$, whereas $\sigma_y^c$ is less influenced by it. Building on this argument, we suggest that the higher accuracy in predicting $E^c$ stems from the fact that it is derived from the initial linear region of the composite’s stress–strain curve, where no nonlinear phenomena (such as yielding) occur, and thus the spatial arrangement of fibres plays no significant role. In contrast, during the onset of yielding, a range of geometric factors beyond $V^{f}$ may affect the composite’s yield behaviour. These include the localised spatial arrangement of fibres, local variations in fibre volume fraction within the microstructure, and even individual inter-fibre distances in critical regions of the microstructure, all of which contribute to the increased complexity in predicting the composite’s yield strength.
Finally, the runtime of the two customised CNNs, i.e., for $E^c$ and $\sigma^c_y$ (including training and testing), was approximately 2 hours, being at least 130 times faster than the total time required for FE simulations (263 hours). This significant reduction in computational cost encourages the application of AI-based approaches in multiscale composite modelling (micro–meso–macro), a challenge that has persisted in composite research for decades. However, in contrast to FE models, which are based on physical laws and explicitly account for the interactions between geometrical and material parameters, CNN models lack interpretability and transparency in their predictive mechanisms. This raises concerns regarding their reliability. In the next section, we look to address this limitation by exploring XAI methods for interpreting the CNN predictions.

\section{Explanation of the CNN Predictions}
\label{sec:xai}

\begin{figure*}[t]
    \centering

    \begin{subfigure}[t]{0.28\textwidth} 
        \centering
        \vbox to 4.8cm{  
            \vfil
            \includegraphics[width=\textwidth]{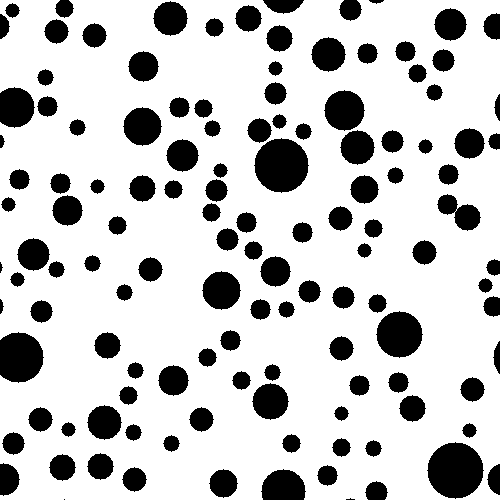}
            \caption{Composite microstructure input image with $V^{f} = 0.22$.}
        }
    \end{subfigure}
    \hfill
    \begin{subfigure}[t]{0.28\textwidth} 
        \centering
        \vbox to 4.8cm{ 
            \vfil
            \includegraphics[width=0.95\textwidth]{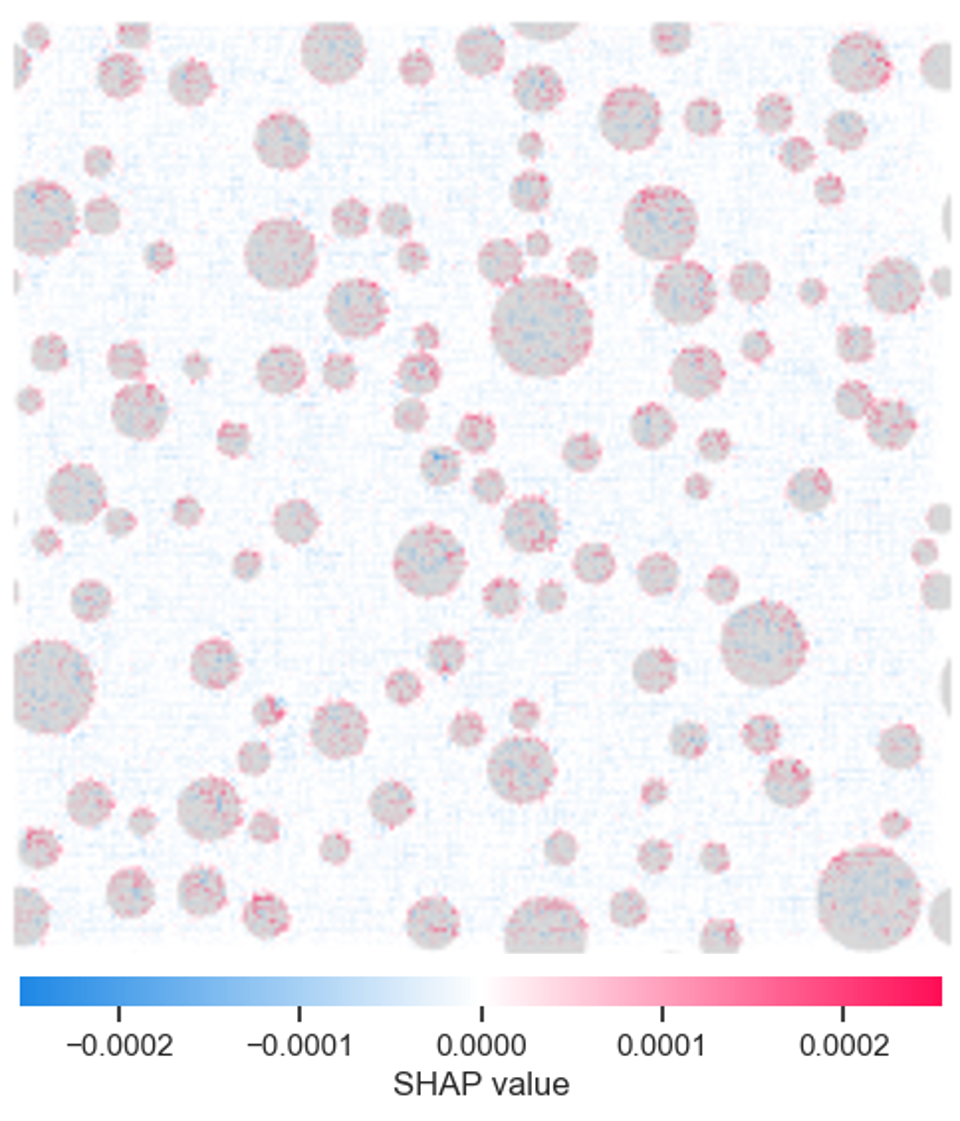}
            \caption{SHAP output for $E^{c}$ for $V^{f} = 0.22$.}
        }
    \end{subfigure}
    \hfill
    \begin{subfigure}[t]{0.28\textwidth} 
        \centering
        \vbox to 4.8cm{ 
            \vfil
            \includegraphics[width=\textwidth]{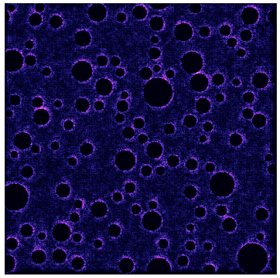}
            \caption{Integrated Gradients output for $E^{c}$ for $V^{f} = 0.22$.}
        }
    \end{subfigure}

    \vspace{1.7cm}

    \begin{subfigure}[t]{0.28\textwidth} 
        \centering
        \vbox to 4.8cm{ 
            \vfil
            \includegraphics[width=\textwidth]{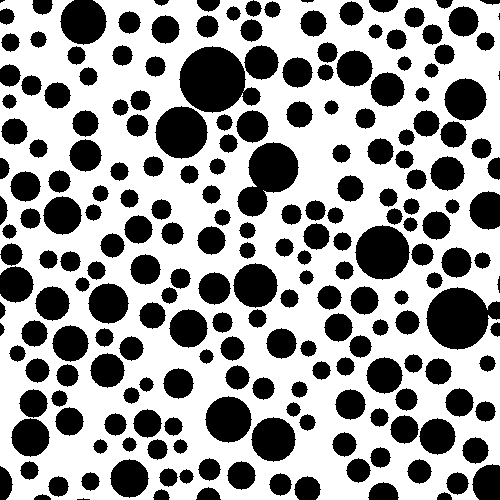}
            \caption{Composite microstructure input image with $V^{f} = 0.4$.}
        }
    \end{subfigure}
    \hfill
    \begin{subfigure}[t]{0.28\textwidth} 
        \centering
        \vbox to 4.8cm{ 
            \vfil
            \includegraphics[width=0.95\textwidth]{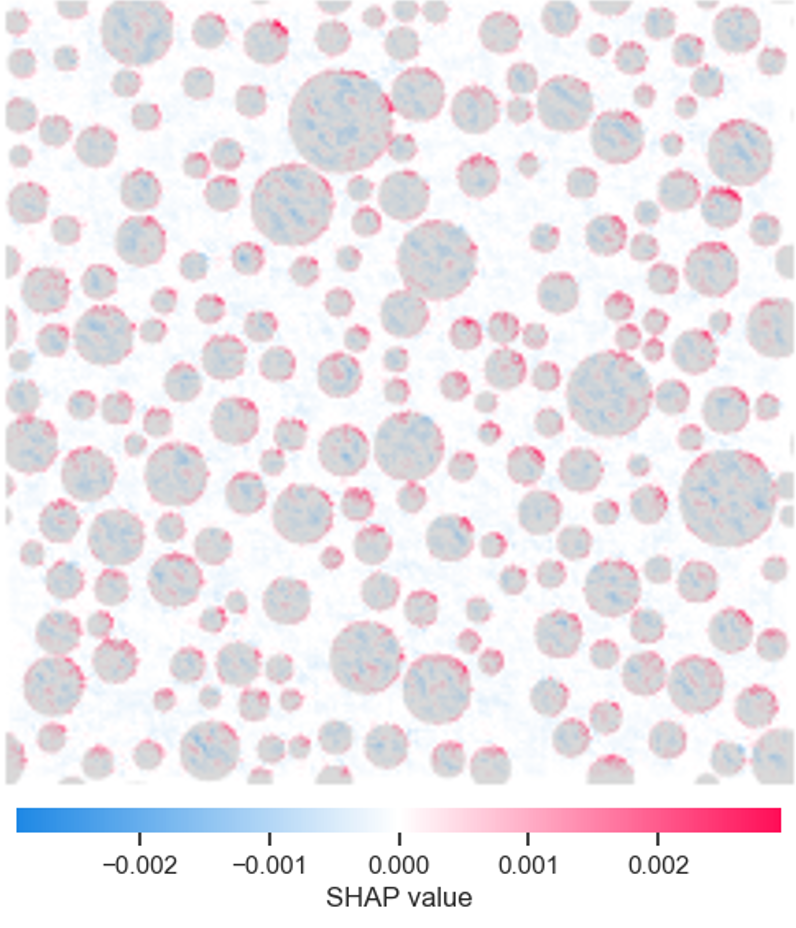}
            \caption{SHAP output for $\sigma_{y}^{c}$ for $V^{f} = 0.4$.}
        }
    \end{subfigure}
    \hfill
    \begin{subfigure}[t]{0.28\textwidth} 
        \centering
        \vbox to 4.8cm{ 
            \vfil
            \includegraphics[width=\textwidth]{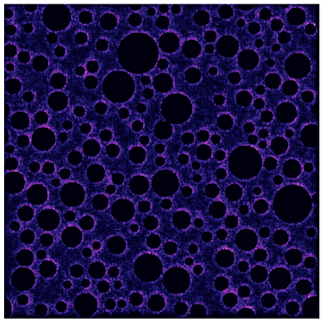}
            \caption{Integrated Gradients output for $\sigma_{y}^{c}$ for $V^{f} = 0.4$.}
        }
    \end{subfigure}
    \vspace{1.6cm}

    \caption{Input images (left column) and corresponding outputs from the XAI methods SHAP (middle column) and Integrated Gradients (right column) for $E^{c}$ with a low volume fraction (top row) and $\sigma_{y}^{c}$ with a high volume fraction (bottom row).}
    \label{fig:xai}
\end{figure*}

We now assess, with a focus on the science of composites, the explanations from two post-hoc XAI methods, SHAP \cite{Lundberg_17} and Integrated Gradients (IG) \cite{Sundararajan_17}, for the predictions made by our customised CNN models. \VR{SHAP was implemented using in-built functions from the SHAP Python library \cite{Lundberg_17}, whilst IG was implemented using the image visualisation method \cite{Sundararajan_17}.}

Figure \ref{fig:xai} presents examples of feature attribution explanations generated by both XAI methods for two representative cases: one with $V^{f} = 0.22$, used to assess $E^{c}$ predictions, and another with $V^{f} = 0.40$, used for $\sigma_{y}^{c}$ predictions. The results from both XAI approaches (SHAP and IG) seem consistent in identifying regions that contribute positively (as shown with red colour for SHAP and magenta for IG) to the predicted output, i.e., $E^{c}$ or $\sigma_{y}^{c}$. Both seem to identify that the regions around the fibres positively contribute to the composite's mechanical properties. Based on these visualisations and the agreement between the two XAI methods, we argue that:

\begin{itemize}
    \item The interface plays a pivotal role in the composite's performance, a fact well established in composite engineering (as depicted by the concentrated positively contributing areas around the fibres). A strong bond at the fibre–matrix interface is known to enhance overall mechanical performance.
    \item The slightly more scattered high-density red regions across the fibre surfaces in the $E^{c}$ assessment case (see Figure \ref{fig:xai}b), with red points appearing not only at the interfaces but also within the fibre interiors, may indicate the direct contribution of the fibres to the enhancement of $E^{c}$, consistent with their role as the primary load-bearing component. As discussed in §\ref{ssec:AIresults}, the strong dependence of $E^{c}$ on the fibre volume fraction ($V^{f}$) may explain the broader spatial distribution of positively contributing regions over the fibre domains.
    \item The SHAP image in Figure~\ref{fig:xai}e reveals a recurring concentration of high-importance regions (red areas) in the top-right quadrant of the fibres across the 2D microstructure. This spatial pattern suggests a potentially critical influence on the predicted $\sigma_{y}^{c}$, possibly corresponding to interface regions that enhance the composite's yield strength, or zones of delayed yielding that contribute positively to the overall yield response. Further validation of this hypothesis would require additional developments in both the FE modelling and our framework, as discussed in §\ref{sec:conclusions}.
\end{itemize}




The fact that only the interface region, and not the fibre surfaces, is highlighted as important w.r.t. $E^c$ and $\sigma_y^c$ by IG, and not SHAP, in Figure \ref{fig:xai}, can be attributed to fundamental differences in the underlying XAI methods. SHAP adopts an approach adapted from cooperative game theory, attributing importance to each pixel in the input image by estimating its marginal contribution to the CNN's output across multiple feature combinations. In contrast, IG quantifies the sensitivity of the CNN's prediction to small changes in the input features by computing the path integral of gradients along a straight line from a predefined baseline image to the input. This may make IG 
particularly sensitive to local variations in pixel intensities, such as those present in the interface region, which may explain the observed differences.










\section{Conclusions and Future Work}
\label{sec:conclusions}

In this paper, we have introduced a novel approach to predicting the mechanical properties, namely $E^c$ and $\sigma^c_y$, of composite materials with CNNs, potentially alleviating the need for computationally expensive and time-consuming FE modelling. 
We propose a customised CNN architecture, adapted to the unorthodox task of image regression, and show it achieves better correlation with FE modelling results than off-the-shelf CNNs, as well as reported results of other AI models from the literature.
Our approach achieved a 99.2\% increase in computational efficiency compared with the massive computational cost of the FE modelling, opening up new possibilities for parametric investigations of composites.
These experiments were performed on a large dataset of composite microstructure images which we have generated and made open-source with the hopes of enabling future research in this area. 
Finally, we have applied feature attribution techniques from XAI, the explanations from which demonstrate that the features deemed important to our trained CNNs aligns with that of the science of composites, potentially helping to foster trust in engineers who may benefit from adopting our approach.

Our findings 
open up many new opportunities in the composite engineering field\DB{, including the following:}
%
%
\begin{itemize}
    \item \DB{Assessing the predictive power of our trained CNNs on real composite microstructures from micro-Computed Tomography tests, which exhibit higher fibre content than the synthetic data, fibre–fibre contact, and local variations in fibre distribution (fibre- or matrix-rich regions).}
    \item 
    Accounting for more advanced constitutive laws than those employed in this study, which more accurately capture the behaviour of composite constituents, i.e., matrix and fibres, would be exceptionally useful, particularly when damage mechanisms are involved. The matrix often exhibits a nonlinear elastic-plastic response with anisotropy (i.e., it does not deform in the same way) across different deformation modes, such as compression, tension, and shear. Incorporating such detail would enable more accurate predictions of composite strength under various loading conditions, which would be extremely beneficial for the assessment of structural integrity. Such an investigation through FE modelling is challenging, computationally expensive (as any sort of nonlinear phenomena adds complexity) and requires high expertise in the selection of appropriate material models. This gives much scope for impact through the application of AI models. 
    \item Accounting for the multiphase nature of composites by explicitly simulating the interface between the fibres and the matrix. The bonding at this interface is often imperfect, which can significantly influence composite performance and lead to failure mechanisms such as fibre-matrix debonding. Capturing this interfacial effect, especially when combined with the use of advanced constitutive laws (as mentioned above), would provide a more comprehensive and accurate framework for predicting the mechanical behaviour of composites. 
    %
    %
    %
    \item Expanding our approach to 3D FE modelling and other loading scenarios, such as longitudinal compression, tension, and shear. Composites exhibit highly anisotropic behaviour, meaning their mechanical response varies significantly depending on the loading direction and fibre orientation. As stated in §\ref{sec:introduction}, 3D FE modelling of composites is computationally expensive for parametric investigations. Thus, an accurate AI model would prove a valuable tool in multiscale composite modelling (micro–meso–macro), a challenge that has drawn attention in composite research for decades.
    \item Finally, employing data-driven insights into composite microstructures through XAI methods can not only increase confidence in the property predictions of AI models, but also help identify critical geometrical or material features that influence composite behaviour, such as yielding, strength, or failure. For example, this framework could reveal localised geometrical characteristics, such as variations in fibre volume fraction or specific fibre arrangements, in critical regions of the 2D microstructure that are likely to initiate yielding or failure. Such an approach could be realised by combining the proposed customised CNN models with XAI techniques, such as SHAP-based superpixel analysis. If such novel insights concerning the physics of composites are discovered via XAI, the impact on composite design, testing and manufacturing could be groundbreaking.
\end{itemize}

\begin{ack}


Bikos kindly acknowledges additional funding for this research provided by the UK Engineering and Physical Sciences Research Council (EPSRC) programme Grant EP/T011653/1, Next Generation Fibre-Reinforced Composites (NextCOMP): a Full-Scale Redesign for Compression in a collaboration between Imperial College London and the University of Bristol. 
Rago and Toni were partially funded by the European Research Council (ERC) under the European Union’s Horizon 2020 research and innovation programme (grant agreement No. 101020934) {and by J.P. Morgan and the Royal Academy of Engineering under the Research Chairs and Senior Research Fellowships scheme (grant agreement no. RCSRF2021\textbackslash 11\textbackslash 45)}.
A Creative Commons Attribution (CC BY-NC 4.0) license is applied to any author-accepted manuscript version arising.

\end{ack}



\bibliography{bib}

\end{document}

%% file: packages.tex
\usepackage{savesym}
\savesymbol{Bbbk}
\savesymbol{openbox}

\usepackage{xcolor}
\usepackage{url}
\usepackage[utf8]{inputenc}
\usepackage{graphicx}
\usepackage{amsmath}
\usepackage{amssymb}
\usepackage{amsthm}
\usepackage{MnSymbol}
\usepackage{booktabs}
\usepackage{algorithm}
\usepackage{algorithmic}
\usepackage{hyperref}
\restoresymbol{NEW}{Bbbk}
\restoresymbol{NEW}{openbox}

%% file: macros.tex
\newcommand{\AR}[1]{\textcolor{black}{#1}}
\newcommand{\DB}[1]{\textcolor{black}{#1}}
\newcommand{\VR}[1]{\textcolor{black}{#1}}
\newcommand{\todo}[1]{\textcolor{magenta}{$*$ #1}}

%% file: main_ArXiv.bbl
\begin{thebibliography}{31}
\providecommand{\natexlab}[1]{#1}
\providecommand{\url}[1]{\texttt{#1}}
\expandafter\ifx\csname urlstyle\endcsname\relax
  \providecommand{\doi}[1]{doi: #1}\else
  \providecommand{\doi}{doi: \begingroup \urlstyle{rm}\Url}\fi

\bibitem[Abueidda et~al.(2019)Abueidda, Almasri, Ammourah, Ravaioli, Jasiuk,
  and Sobh]{abueidda2019}
D.~W. Abueidda, M.~Almasri, R.~Ammourah, U.~Ravaioli, I.~M. Jasiuk, and N.~A.
  Sobh.
\newblock Prediction and optimization of mechanical properties of composites
  using convolutional neural networks.
\newblock \emph{Composite Structures}, 227:\penalty0 111264, 2019.
\newblock URL \url{https://doi.org/10.1016/j.compstruct.2019.111264}.

\bibitem[Ali et~al.(2023)Ali, Abuhmed, El{-}Sappagh, Muhammad, Alonso{-}Moral,
  Confalonieri, Guidotti, Ser, Rodr{\'{\i}}guez, and Herrera]{Ali_23}
S.~Ali, T.~Abuhmed, S.~H.~A. El{-}Sappagh, K.~Muhammad, J.~M. Alonso{-}Moral,
  R.~Confalonieri, R.~Guidotti, J.~D. Ser, N.~D. Rodr{\'{\i}}guez, and
  F.~Herrera.
\newblock Explainable artificial intelligence {(XAI):} what we know and what is
  left to attain trustworthy artificial intelligence.
\newblock \emph{Inf. Fusion}, 99:\penalty0 101805, 2023.
\newblock URL \url{https://doi.org/10.1016/j.inffus.2023.101805}.

\bibitem[Bikos et~al.(2025)Bikos, Poh, Trask, Robinson, and Pimenta]{bikos2025}
D.~Bikos, F.~Poh, R.~Trask, P.~Robinson, and S.~Pimenta.
\newblock Shell-beam micromechanical models to improve the efficiency of
  simulations of composites under longitudinal compression.
\newblock \emph{Composite Structures}, 354:\penalty0 118830, 2025.
\newblock URL \url{https://doi.org/10.1016/j.compstruct.2024.118830}.

\bibitem[Chen et~al.(2021)Chen, Wan, Ismail, Ye, and Yang]{chen2021}
J.~Chen, L.~Wan, Y.~Ismail, J.~Ye, and D.~Yang.
\newblock A micromechanics and machine learning coupled approach for failure
  prediction of unidirectional cfrp composites under triaxial loading: A
  preliminary study.
\newblock \emph{Composite Structures}, 267:\penalty0 113876, 2021.
\newblock URL \url{https://doi.org/10.1016/j.compstruct.2021.113876}.

\bibitem[Chen and Guestrin(2016)]{Chen_16}
T.~Chen and C.~Guestrin.
\newblock {XGB}oost: {A} scalable tree boosting system.
\newblock In \emph{SIGKDD}, pages 785--794, 2016.
\newblock \doi{10.1145/2939672.2939785}.
\newblock URL \url{https://doi.org/10.1145/2939672.2939785}.

\bibitem[Ford et~al.(2021)Ford, Maneparambil, Rajan, and Neithalath]{Ford_21}
E.~Ford, K.~Maneparambil, S.~Rajan, and N.~Neithalath.
\newblock Machine learning-based accelerated property prediction of two-phase
  materials using microstructural descriptors and finite element analysis.
\newblock \emph{Computational Materials Science}, 191:\penalty0 110328, 2021.
\newblock URL \url{https://doi.org/10.1016/j.commatsci.2021.110328}.

\bibitem[Harper et~al.(1993)Harper, Miller, and Yap]{harper1993}
J.~Harper, N.~Miller, and S.~Yap.
\newblock Problems associated with the compression testing of fibre reinforced
  plastic composites.
\newblock \emph{Polymer testing}, 12\penalty0 (1):\penalty0 15--29, 1993.
\newblock URL \url{https://doi.org/10.1016/0142-9418(93)90023-I}.

\bibitem[He and Gao(2015)]{he2015}
H.-w. He and F.~Gao.
\newblock Effect of fiber volume fraction on the flexural properties of
  unidirectional carbon fiber/epoxy composites.
\newblock \emph{International Journal of Polymer Analysis and
  Characterization}, 20\penalty0 (2):\penalty0 180--189, 2015.
\newblock URL \url{https://doi.org/10.1080/1023666X.2015.989076}.

\bibitem[He et~al.(2016)He, Zhang, Ren, and Sun]{He_16}
K.~He, X.~Zhang, S.~Ren, and J.~Sun.
\newblock Deep residual learning for image recognition.
\newblock In \emph{{CVPR}}, pages 770--778, 2016.
\newblock URL \url{https://doi.org/10.1109/CVPR.2016.90}.

\bibitem[{Joshua Robbins}(2019)]{joshua_robbins_building_2019}
{Joshua Robbins}.
\newblock On {Building} {Blocks} for {Virtual} {Testing} of {Unidirectional}
  {Polymeric} {Composites}, Dec. 2019.
\newblock URL
  \url{https://keep.lib.asu.edu/system/files/c7/220493/Robbins_asu_0010N_19564.pdf}.

\bibitem[Larson et~al.(2022)Larson, Hoque, Jamora, Li, Kravchenko, and
  Kravchenko]{larson2022}
R.~A. Larson, R.~Hoque, V.~Jamora, J.~Li, S.~Kravchenko, and O.~Kravchenko.
\newblock Hyperparameters effect in deep convolutional neural network model on
  prediction of fiber orientation distribution in prepreg platelet molded
  composites.
\newblock In \emph{AIAA SCITECH 2022}, page 0103, 2022.
\newblock URL \url{https://doi.org/10.2514/6.2022-0103}.

\bibitem[Li et~al.(2022)Li, Liu, Yang, Peng, and Zhou]{Li_22}
Z.~Li, F.~Liu, W.~Yang, S.~Peng, and J.~Zhou.
\newblock A survey of convolutional neural networks: Analysis, applications,
  and prospects.
\newblock \emph{{IEEE} Trans. Neural Networks Learn. Syst.}, 33\penalty0
  (12):\penalty0 6999--7019, 2022.
\newblock URL \url{https://doi.org/10.1109/TNNLS.2021.3084827}.

\bibitem[Lundberg and Lee(2017)]{Lundberg_17}
S.~M. Lundberg and S.~Lee.
\newblock A unified approach to interpreting model predictions.
\newblock In \emph{NeurIPS}, pages 4765--4774, 2017.
\newblock URL
  \url{https://proceedings.neurips.cc/paper/2017/hash/8a20a8621978632d76c43dfd28b67767-Abstract.html}.

\bibitem[Malgioglio et~al.(2021)Malgioglio, Pimenta, Matveeva, Farkas, Desmet,
  Lomov, and Swolfs]{malgioglio2021}
F.~Malgioglio, S.~Pimenta, A.~Matveeva, L.~Farkas, W.~Desmet, S.~V. Lomov, and
  Y.~Swolfs.
\newblock Microscale material variability and its effect on longitudinal
  tensile failure of unidirectional carbon fibre composites.
\newblock \emph{Composite structures}, 261:\penalty0 113300, 2021.
\newblock URL \url{https://doi.org/10.1016/j.compstruct.2020.113300}.

\bibitem[Pathan et~al.(2019)Pathan, Ponnusami, Pathan, Pitisongsawat, Erice,
  Petrinic, and Tagarielli]{pathan2019}
M.~Pathan, S.~Ponnusami, J.~Pathan, R.~Pitisongsawat, B.~Erice, N.~Petrinic,
  and V.~Tagarielli.
\newblock Predictions of the mechanical properties of unidirectional fibre
  composites by supervised machine learning.
\newblock \emph{Scientific reports}, 9\penalty0 (1):\penalty0 13964, 2019.
\newblock URL \url{https://doi.org/10.1038/s41598-019-50144-w}.

\bibitem[Post et~al.(2023)Post, Lin, Waas, and Ustun]{post2023}
A.~Post, S.~Lin, A.~M. Waas, and I.~Ustun.
\newblock Determining damage initiation of carbon fiber reinforced polymer
  composites using machine learning.
\newblock \emph{Polymer Composites}, 44\penalty0 (2):\penalty0 932--953, 2023.
\newblock URL \url{https://doi.org/10.1002/pc.27144}.

\bibitem[Qiu et~al.(2022)Qiu, Han, Shanmugam, Zhao, Dong, Du, and
  Yang]{qiu2022}
C.~Qiu, Y.~Han, L.~Shanmugam, Y.~Zhao, S.~Dong, S.~Du, and J.~Yang.
\newblock A deep learning-based composite design strategy for efficient
  selection of material and layup sequences from a given database.
\newblock \emph{Composites Science and Technology}, 230:\penalty0 109154, 2022.
\newblock URL \url{https://doi.org/10.1016/j.compscitech.2021.109154}.

\bibitem[Rajak et~al.(2021)Rajak, Wagh, and Linul]{rajak2021}
D.~K. Rajak, P.~H. Wagh, and E.~Linul.
\newblock Manufacturing technologies of carbon/glass fiber-reinforced polymer
  composites and their properties: A review.
\newblock \emph{Polymers}, 13\penalty0 (21):\penalty0 3721, 2021.
\newblock URL \url{https://doi.org/10.3390/polym13213721}.

\bibitem[Raju et~al.(2018)Raju, Hiremath, and Mahapatra]{raju2018}
B.~Raju, S.~Hiremath, and D.~R. Mahapatra.
\newblock A review of micromechanics based models for effective elastic
  properties of reinforced polymer matrix composites.
\newblock \emph{Composite Structures}, 204:\penalty0 607--619, 2018.
\newblock URL \url{https://doi.org/10.1016/j.compstruct.2018.07.125}.

\bibitem[Samaras et~al.(2024)Samaras, Bikos, Cann, Masen, Hardalupas, Vieira,
  Hartmann, and Charalambides]{samaras2024}
G.~Samaras, D.~Bikos, P.~Cann, M.~Masen, Y.~Hardalupas, J.~Vieira, C.~Hartmann,
  and M.~Charalambides.
\newblock A multiscale finite element analysis model for predicting the effect
  of micro-aeration on the fragmentation of chocolate during the first bite.
\newblock \emph{European Journal of Mechanics-A/Solids}, 104:\penalty0 105221,
  2024.
\newblock URL \url{https://doi.org/10.1016/j.euromechsol.2024.105221}.

\bibitem[Saquib et~al.(2024)Saquib, Larson, Sattar, Li, Kravchenko, and
  Kravchenko]{saquib2024}
M.~N. Saquib, R.~Larson, S.~Sattar, J.~Li, S.~G. Kravchenko, and O.~G.
  Kravchenko.
\newblock Experimental validation of reconstructed microstructure via deep
  learning in discontinuous fiber platelet composite.
\newblock \emph{Journal of Applied Mechanics}, 91\penalty0 (4):\penalty0
  041004, 2024.
\newblock URL \url{https://doi.org/10.1115/1.4063983}.

\bibitem[Smith(2009)]{Abaqus}
M.~Smith.
\newblock \emph{ABAQUS/Standard User's Manual, Version 6.9}.
\newblock Dassault Syst{\`e}mes Simulia Corp, United States, 2009.
\newblock URL
  \url{https://classes.engineering.wustl.edu/2009/spring/mase5513/abaqus/docs/v6.6/books/usb/default.htm}.

\bibitem[Sundararajan et~al.(2017)Sundararajan, Taly, and Yan]{Sundararajan_17}
M.~Sundararajan, A.~Taly, and Q.~Yan.
\newblock Axiomatic attribution for deep networks.
\newblock In \emph{{ICML}}, pages 3319--3328, 2017.
\newblock URL \url{http://proceedings.mlr.press/v70/sundararajan17a.html}.

\bibitem[Tschopp et~al.(2008)Tschopp, Wilks, and Spowart]{tschopp2008}
M.~Tschopp, G.~Wilks, and J.~Spowart.
\newblock Multi-scale characterization of orthotropic microstructures.
\newblock \emph{Modelling and Simulation in Materials Science and Engineering},
  16\penalty0 (6):\penalty0 065009, 2008.
\newblock URL \url{https://dx.doi.org/10.1088/0965-0393/16/6/065009}.

\bibitem[Varandas et~al.(2020)Varandas, Catalanotti, Melro, Tavares, and
  Falzon]{varandas2020}
L.~F. Varandas, G.~Catalanotti, A.~R. Melro, R.~Tavares, and B.~G. Falzon.
\newblock Micromechanical modelling of the longitudinal compressive and tensile
  failure of unidirectional composites: The effect of fibre misalignment
  introduced via a stochastic process.
\newblock \emph{International journal of solids and structures}, 203:\penalty0
  157--176, 2020.
\newblock URL \url{https://doi.org/10.1016/j.ijsolstr.2020.07.022}.

\bibitem[Wan et~al.(2023)Wan, Ullah, Yang, and Falzon]{wan2023}
L.~Wan, Z.~Ullah, D.~Yang, and B.~G. Falzon.
\newblock Probability embedded failure prediction of unidirectional composites
  under biaxial loadings combining machine learning and micromechanical
  modelling.
\newblock \emph{Composite Structures}, 312:\penalty0 116837, 2023.
\newblock URL \url{https://doi.org/10.1016/j.compstruct.2023.116837}.

\bibitem[Wang et~al.(2019)Wang, Chai, Soutis, and Withers]{wang2019}
Y.~Wang, Y.~Chai, C.~Soutis, and P.~J. Withers.
\newblock Evolution of kink bands in a notched unidirectional carbon
  fibre-epoxy composite under four-point bending.
\newblock \emph{Composites Science and Technology}, 172:\penalty0 143--152,
  2019.
\newblock URL \url{https://doi.org/10.1016/j.compscitech.2019.01.014}.

\bibitem[Wang et~al.(2024)Wang, Wang, and Zhang]{wang2024}
Y.~Wang, K.~Wang, and C.~Zhang.
\newblock Applications of artificial intelligence/machine learning to
  high-performance composites.
\newblock \emph{Composites Part B: Engineering}, page 111740, 2024.
\newblock URL \url{https://doi.org/10.1016/j.compositesb.2024.111740}.

\bibitem[Yiben et~al.(2024)Yiben, Guangshuo, and Bo]{yiben2024}
Z.~Yiben, F.~Guangshuo, and L.~Bo.
\newblock A machine learning based prediction model for the impact mechanical
  response of composite laminates considering microstructure sensitive
  transverse properties.
\newblock \emph{Polymer Composites}, 2024.
\newblock URL \url{https://doi.org/10.1002/pc.29203}.

\bibitem[Yossef et~al.(2024)Yossef, Noureldin, and Alqabbany]{yossef2024}
M.~Yossef, M.~Noureldin, and A.~Alqabbany.
\newblock Explainable artificial intelligence framework for frp composites
  design.
\newblock \emph{Composite Structures}, 341:\penalty0 118190, 2024.
\newblock URL \url{https://doi.org/10.1016/j.compstruct.2024.118190}.

\bibitem[Zhou et~al.(2021)Zhou, Sun, Enos, Zhang, and Tang]{zhou2021}
K.~Zhou, H.~Sun, R.~Enos, D.~Zhang, and J.~Tang.
\newblock Harnessing deep learning for physics-informed prediction of composite
  strength with microstructural uncertainties.
\newblock \emph{Computational Materials Science}, 197:\penalty0 110663, 2021.
\newblock URL \url{https://doi.org/10.1016/j.commatsci.2021.110663}.

\end{thebibliography}
